# A Cooperative Dynamic Task Assignment Framework for COTSBot AUVs

Amin Abbasi, Somaiyeh MahmoudZadeh, Amirmehdi Yazdani

*Abstract*—This paper presents a cooperative dynamic task assignment framework for a certain class of Autonomous Underwater Vehicles (AUVs) employed to control outbreak of Crown-Of-Thorns Starfish (COTS) in the Australia's Great Barrier Reef. The problem of monitoring and controlling the COTS is transcribed into a constrained task assignment problem in which eradicating clusters of COTS, by the injection system of COTSbot AUVs, is considered as a task. A probabilistic map of the operating environment including seabed terrain, clusters of COTS, and coastlines is constructed. Then, a novel heuristic algorithm called Heuristic Fleet Cooperation (HFC) is developed to provide cooperative injection of the COTSbot AUVs to the maximum possible COTS in an assigned mission time. Extensive simulation studies together with quantitative performance analysis are conducted to demonstrate the effectiveness and robustness of the proposed cooperative task assignment algorithm in eradicating the COTS in the Great Barrier Reef.

*Note to Practitioners*—This research is motivated by controlling outbreak of COTS, in the Australia's Great Barrier Reef, by employing a certain class of AUVs equipped with the injection arm. A new evolutionary algorithm, called HFC, is designed upon a hierarchal priority evolution to facilitate multivehicle dynamic task assignment problem. The HFC is able to rapidly prototype multiple optimal solutions for efficient grouping and accomplishment of the tasks spread over a large operation area.

*Index Terms*— Cooperation, COTSbot AUVs, Crown-Of-Thorns Starfish, Task assignment

## I. Introduction

IN recent years, autonomous underwater vehicles (AUVs) have been used excessively in various ranges of oceanic surveys and applications such as mine detection, offshore infrastructure manipulation and maintenance, mapping, data collection and sampling, and monitoring [1], [2]. For example, for environmental scientists, an AUV can be a great tool to collect submerged data of marine organisms such as sediment transport in degraded ecosystems [3], [4].

Due to the unique underwater ecological nature of Australia, monitoring undersea terrains and biological organisms have been always important for Australian Government and environmentalists. One of the challenging issues of recent years that the Queensland's Government has been dealt with is the destructive impact of COTS destroyed more than 50% of the coral in the Great Barrier Reef [5]. Different measures have been taken into account for this particular problem, however, using traditional approaches in removing starfish, undertaken with human divers, are not operationally effective and logistically economical. Thus, Queensland University of Technology (QUT) was developed a particular class of AUV called COTSbot AUV equipped with vison-based technology and an injection arm to eradicate COTS in the Great Barrier Reef automatically [6]. Even though the employment of the proposed AUV was promising and effective in protection of corals against population of COTS compared to the traditional approaches, however, this performance can be further improved by employing a group of COTSbot AUVs working cooperatively in the Great Barrier Reef region. The main purpose of this paper is to use COTSbot AUVs in the context of multi-robot task assignment (MRTA) to enhance the operational performance in conjunction with saving operation cost and time for the particular problem of control of COTS in the Great Barrier Reef.

MRTA systems are usually utilized to perform missions which are time-consuming or difficult to be carried out by a single robot/vehicle as they are more capable of handling large-scale and complex missions and more fault-tolerant than a single robot system [7]. The main objectives of using MRTA technology is to maximize the operational performance and to minimize operation time/cost; these are achieved by organizing the underlying mission in the context of task allocation and assignment in which clustering, classifying, prioritizing, and accomplishment of tasks are key elements. For example, in [8], a group of Unmanned Aerial Vehicles (UAVs) was used in a detect-and-treat mission to first identify the palm trees which infested by weevils and then treat them by pesticide. The mission was defined as an MRTA problem, and a bio-inspired algorithm based on bacteria foraging behavior was employed to solve the problem. In [9], a search-and-rescue mission was carried out by multiple UAVs in which the performance impact algorithm was proposed to define a dynamic grouping allocation to deal with communication disruptions and to avoid conflict in task assignment of UAVs. In [10], a MRTA problem was defined for a team of AUVs in which a workload balance algorithm was developed for investigation of suspicious objects in an underwater environment in presence of obstacles and wave disturbance. In [11], an MRTA problem which considers the time utility and energy consumption of a team of robot was mathematically formulated as a multi-objective optimization problem and solved by multi-objective PSO algorithm. In [12], a team of wheeled mobile robots was employed to carry items from storage racks to packing dock in a factory. The items had

Amin Abbasi is with the Department of Electrical Engineering, Azad University of Khoemeinishar, Esfahan, Iran (e-mail: aminabbasi.res@gmail.com)

Somaiyeh MahmoudZadeh is with the School of IT, Deakin University, Geelong, VIC 3220, Australia (e-mail: s.mahmoudzadeh@deakin.edu.au)

Amirmehdi Yazdani is with the College of Science, Health, Engineering and Education, Murdoch University, Perth, WA 6150, Australia (e-mail: amirmehdi.yazdani@murdoch.edu.au)

different weights and all the robots had similar maximum weight capacity. The nearest-neighbor algorithm based on clustering and routing methods was used to minimize the total distance and fetch as many items as possible within the weight capacity limit. In [13], a cooperative package delivery mission was studied and transformed into the context of multi-agent task assignment problem. Micro drone vehicles, carried and supported by a moving truck, were employed as delivery agents. Operating range and load capacity of the drones as well as the street network restrictions were introduced to the problem as main constraints. The customers were prioritized by their urgent priority level, and the main purpose was to deliver all packages within the minimum possible time along the truck's route. Implementing a variety of scenarios under different testing conditions revealed that the proposed algorithm is more efficient in comparison with conventional genetic algorithm. In [14], a multi-agent task assignment algorithm was developed for a multi unmanned surface vehicle (multi-USV) system. Using artificial neural networks, a low dimensional representation of the working environment, named self-organizing map was utilized for task allocation process and dividing the area among the USVs. To plan the optimal path for the vehicles, a trajectory planning method based on fast marching method was proposed to accomplish the determined mission considering the energy consumption, communication range, and collision avoidance. The results of this study indicate the effectiveness of the algorithm in both simulation and real maritime operating field. In [15], a multi-AUV collaborative system was employed for target search and tracking. Combination of the Glasius Bio-inspired Neural Network (GBNN) and a cascade tracking control was used to navigate a multi-AUV collaborative system. The GBNN was utilized to establish environmental perception, and the cascade control was responsible for target tracking task. The proposed system was able to reduce the tracking error and worked in dynamic environments cooperatively. In [16], a task assignment problem was defined for a multi-UAV system to attack dynamic targets in a battlefield. To solve the problem, the combination of two algorithms was utilized: a dynamic programming algorithm was established for route planning of the vehicles, and a multi-subgroup ant colony algorithm was proposed to optimally assign the task to the UAVs. The main contributing factor of this study is to plan feasible flying routes compatible with the dynamic of aerial vehicles, and to consider moving targets instead of fixed targets in the mission. In [17], a self-organizing map (SOM) algorithm was modified by a biologically-inspired neural network (BINN) for task assignment and path planning of a multi-AUV system. The SOM algorithm was responsible for allocating the tasks to the vehicles, and the BINN algorithm was in charge of generating a collision-free path considering the saturation level of the vehicles' speed. However, in this study, the environmental uncertainties and disturbances were not taken into consideration. Finally, in [18], a dynamic task assignment and path planning framework was defined for a group of AUVs. The algorithm was an integration of SOM, neural network, and a mechanism for velocity synthesis. The proposed framework enabled the multi-AUV system to reach several appointed targets in presence of ocean current while considering the workload balance and energy consumption.

The main contribution of this paper is to develop a new Heuristic Fleet Cooperation (HFC) algorithm as a population-based evolutionary approach for solving multi-vehicle dynamic task assignment problem, where the framework specifically designed for COTSbot AUVs to effectively accommodate the problem of control of COTS in the Queensland's Great Barrier Reef. The HFC algorithm is established upon a hierarchal priority evolution and is able to rapidly prototype multiple optimal solutions for efficient grouping and accomplishment of the existing tasks spread over a large operation area. The detailed mechanism of the algorithm is explained in Section III.

To the best of authors' knowledge, this is the first study of using a team of COTSbot AUVs for eradicating the COTS in the Great Barrier Reef. To this end, the problem of controlling COTS is transcribed into a cooperative task assignment problem in which eradicating the COST via the injection system mounted on the AUVs is defined as a task. A novel cooperative task assignment algorithm including Task Clustering, Ordering, Screening, and Cooperation operators are developed to provide cooperative injection of the COTSbot AUVs to the maximum possible COTS in the assigned mission time. Extensive simulation studies including the comparative performance assessment together with the quantitative performance analysis are conducted to demonstrate the effectiveness and robustness of the proposed cooperative task assignment algorithm in eradicating the COTS in the Great Barrier Reef.

The rest of this paper is organized as follows. Section 2 provides a clear picture of problem formulation including details of the COTS control mission, the probabilistic map of the operating field, and mathematical model of the AUV. In Section 3, details of the COTS-based cooperative task assignment algorithm are presented. Section 4 shows extensive simulation studies and performance evaluation of the proposed algorithm in a COTS control mission. Finally, Section 5, concludes the paper.

## II. Problem Formulation

Figure 1 illustrates the concept of employing multiple COTSbot AUVs for COTS control. The COTSbot AUV is equipped with a robotic arm and injection mechanism with an onboard vision-based controller [6] to coordinate arm movement and vehicles position according to the COTS location. A partial map of the Barrier Reef region with the Latitude of ⟨22.32° S 143.09° E⟩ to ⟨9.34° S 143.46°E⟩ and longitude of ⟨22.32° S 143.09° E⟩ to ⟨22.5° S 157.5° E⟩ is provided for the AUVs operation, presented in Fig.2.

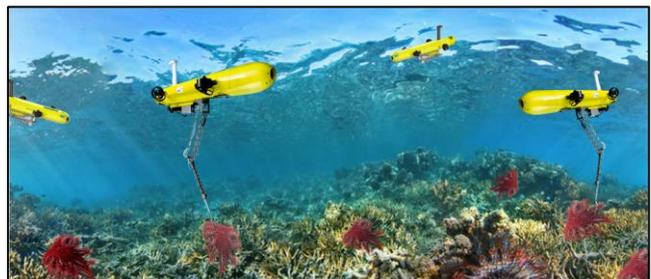

Fig. 1. Concept of employing multiple AUVs equipped with the injection system for COTS control.

Multiple vehicles' cooperative operations would be a useful idea to minimize the cost of deployment, launch and recovery, and to increase the efficiency of the undersea missions restricted by a vehicle battery capacity. This research aims to use the COTSBot AUVs to identify COTS within complex reef environments and perform injection to eradicate them. The AUVs should cooperatively inject the maximum possible COTS in the assigned mission time. The vehicles should be able to update each other about the COTS areas cleaned up to increase the effectiveness of coverage and to avoid duplicating the treatment.

*Remark* 1- The AUVs are equipped with the acoustic navigation aids such as digital ultra-short baseline (DUSBL), and therefore they are able to share their localization information and coordinate of COTS areas to each other in a fixed sample time via the acoustic communication. This share/exchange of information between the AUVs contributes to effectiveness of coverage and avoiding duplicating the treatment.

*Assumption* 1- It is assumed that the AUVs use the constant thrust power during the mission and therefore the average vehicles' velocity is constant.

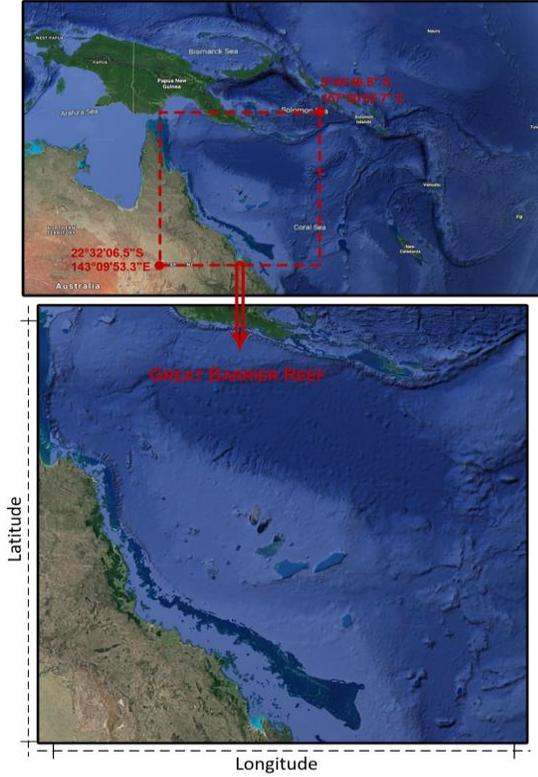

Fig. 2. A snapshot of selected map area in the Queensland's Great Barrier Reef region with the Latitude of ⟨22.32° S 143.09° E⟩ to ⟨9.34° S 143.46°E⟩ and longitude of ⟨22.32° S 143.09° E⟩ to ⟨22.5° S 157.5° E⟩.

The problem of controlling COTS is transcribed into a cooperative task assignment problem in which eradicating the COST (via the injection system mounted on the AUVs) is defined as a task. The AUVs are given a probabilistic map of the undersea environment prior to the mission, and they use the developed vison-based technology [6], to accurately distinguish the COTS and to perform the injection process. During the mission, each vehicle should exchange the information of its pose, COTS coordinate, and number of tasks completed with other vehicle. In a case that a vehicle aborts its mission for any reason (e.g., ran out of battery), the closest vehicle undertakes the incomplete mission, or the mission is rearranged between several vehicles and a new mission scenario for all vehicles is planned. In the subsequent sections, the detailed mathematical representations of this problem are provided.

### A. Modelling the operation field and distribution of the COTS

A prior knowledge of the terrain such as coastal areas, forbidden operation zones, and the coordinate of start and endpoint enhances AUVs' capability in robust motion planning. Even though preparing a perfect offline map is rarely possible in undersea operations, AUVs can take advantage of any partially constructed map to have a rough perception of the operating field in a priori. The seabed terrain is modelled using a numerical estimated model of the field, derived from the following equations shown in (1):

$$\tau_{x,y}^{\mathbb{V}} = \begin{cases} \tau_x^{\mathbb{V}} = \frac{(\mathfrak{T}y - \mathfrak{T}y_{\mathcal{O}})\left(e^{-(\mathbb{V}-\mathcal{O})^2 r^{-2}} - 1\right)}{2\pi(\mathbb{V}-\mathcal{O})^2} \\ \tau_y^{\mathbb{V}} = \frac{(\mathfrak{T}x - \mathfrak{T}x_{\mathcal{O}})\left(1 - e^{-(\mathbb{V}-\mathcal{O})^2 r^{-2}}\right)}{2\pi(\mathbb{V}-\mathcal{O})^2} \end{cases} \quad (1)$$

where, $\tau_{x,y}$ represents non-uniform distributions of COTS in a 2D plane of volume of $\mathbb{V}$; $\mathcal{O}$ and $r$ are the centre and radius of high-density COTS area; $\mathfrak{T}$ corresponds to density of distribution around each centre ($\mathcal{O}$). Information of coastlines and islands locations are known priori and included in the map. Probability of COTS distribution in the $\mathbb{V}_{2D}:\langle\{1 \times 1\ km^2\}_{x-y}\rangle$ area is depicted by Fig.3.

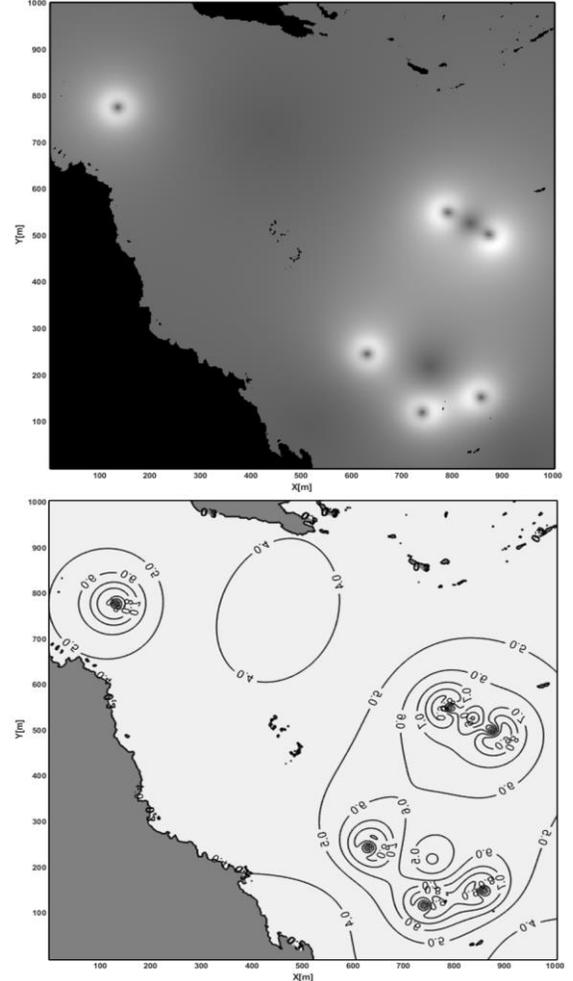

Fig. 3. Probability of COTS distribution in the operation environment captured from Australia's Great Barrier Reef.

A set of AUVs can be used to detect COTs agglomeration and provide tracking information to the robot's arm to inject the COTs. Therefore, the tasks in this work are defined as injecting the COTs, where each task associated with a priority rank and completion time depending on the intensity of COTs in a spot. The AUVs are required to consistently find COTs within complex reef environments, and cooperatively eradicate maximum number of COTs while moving toward the destination. Therefore, tasks should be arranged to govern the AUVs to the destination and meet the time restriction, which is a joint discrete and continuous multi-objective problem resembling both Knapsack and TSP problems. Concentration of the COTs is presented with $\mathcal{C} = \{C_1, \ldots, C_j, \ldots, C_n\}$, where each node $C_j$ is a spot to be injected. In this context, the AUV mission planner simultaneously tends to determine the optimum order of COTS spots to be injected mathematically described as follows:

$$\mathcal{C} = \{C_1, \ldots, C_j, \ldots, C_n\};$$
$$\forall C_j, \exists \rho^{C_j} \sim \mathbb{U}(1,100), \quad (2)$$
$$\forall C_j, \exists t^{C_j} \sim \mathbb{U}(60,90)$$

where, $t^{C_j}$ is the time required to complete injection, which depends on COTs density in the area, and can range between 60 to 90 seconds. $\rho^{C_j}$ is the task priority rank. The high intensity areas are associated with a greater number of tasks (killing COTs) which should be completed.

### B. Mathematical Model of Multiple Vehicles Operation

Assuming there are $k$ number of identical AUVs in the fleet $\mathcal{A} = \langle \mathcal{A}_1, \ldots, \mathcal{A}_k \rangle$ with six degrees of freedom for translational and rotational motion in NED $\{n\}$ and Body $\{b\}$ frames; the physical model of $\mathcal{A}_i$ moving in a 3D volume is described as follows [19, 20]:

$$\forall \mathcal{A}_i, \begin{array}{l} \{n\} \to \exists \eta_i = [x_i, y_i, z_i, \varphi_i, \theta_i, \psi_i]^T \\ \{b\} \to \exists v_i = [v_{i,x}, v_{i,y}, v_{i,z}, p_i, q_i, r_i]^T \end{array} \quad (3)$$

$$\begin{cases} v_{i,x} = |v_i| \cos\theta_i \cos\psi_i \\ v_{i,y} = |v_i| \cos\theta_i \sin\psi_i \\ v_{i,z} = |v_i| \sin\theta_i \end{cases} \quad (4)$$

where, $\eta_i$ denotes the $\mathcal{A}_i$ state vector on NED, including the position in North, $x_i$, East, $y_i$, Down, $z_i$ and the Euler angles of roll $\varphi_i$, pitch $\theta_i$ and yaw $\psi_i$ motions. The $v_i: \langle v_{i,x}, v_{i,y}, v_{i,z} \rangle$ is the $\mathcal{A}_i$ translational velocity vector along the surge, sway and heave directions; while $v_i: \langle p_i, q_i, r_i \rangle$ is the $\mathcal{A}_i$ vector of rotational velocity. The vehicle rotation along the z-axis (yaw angle) and y-axis (pitch angle) are obtained via (5)–(6).

$$\psi_i(t) = \arctan\left(\frac{\Delta y_i(t)}{\Delta x_i(t)}\right) \quad (5)$$

$$\theta_i(t) = \arctan\left(\frac{-\Delta z_i(t)}{\sqrt{(\Delta y_i(t))^2 + (\Delta x_i(t))^2}}\right) \quad (6)$$

*Assumption* 2- The vehicle rotation along the *x*-axis, roll angle, is assumed to be negligible in this study.

The distance travelled by vehicle $\mathcal{A}_i$ from spot $C_j$ to $C_l$ is calculated via (7).

$$\mathcal{D}_{\mathcal{A}_i}^{C_{j,l}}(t) = \sqrt{\left(\Delta x_{\mathcal{A}_i}(t)\right)^2 + \left(\Delta y_{\mathcal{A}_i}(t)\right)^2 + \left(\Delta z_{\mathcal{A}_i}(t)\right)^2} \quad (7)$$

The objective of multi-AUVs mission planning system is to find an optimal route $\Re$ that maximizes the total number of injected COTs (in a possible widest area) for a restricted mission time for each vehicle ($\mathcal{T}_{\mathcal{A}_i}^{\nabla}$) while optimizing some performance indices such as operation time and travel distance. To this end, the vehicles move through the high-density COTS areas following the route generated by the mission planning system, where on-time visit to the target station is the main concern of the framework. Due to energy restrictions and extensive number of COTS distributed in a large operation field, completing all tasks in one mission is not feasible for a limited number of vehicles. Therefore, an impact factor of $\rho$ has been assigned to COTS centers to prioritize the order of tasks which should be completed and govern the vehicles toward the destination. In this framework, any arbitrary route $\Re$ traveled by $\mathcal{A}_i$ is characterized by the corresponding time $\mathcal{T}_{\mathcal{A}_i}^M$ required for travelling $\Re_{\mathcal{A}_i}$ and completing injection process on each $C_j$, which is modelled by:

$$\Re_{\mathcal{A}_i}: \langle S_{xyz}^{s_i}, \ldots, C_{j,x,yz}, C_{l,x,yz}, \ldots, S_{xyz}^{G_i} \rangle$$
$$\forall, C_{l,x,yz}; \exists \left(\mathcal{D}_{C_{j,l}}, \mathcal{T}_{C_{j,l}}, \rho_{C_j}, \rho_{C_l}\right)$$
$$\mathcal{D}_{C_{j,l}} = \sqrt{(C_{l,x} - C_{j,x})^2 + (C_{l,y} - C_{j,y})^2 + (C_{l,z} - C_{j,z})^2} \quad (8)$$
$$\forall \mathcal{D}_{C_{j,l}}; \exists \mathcal{T}_{C_{j,l}} = \mathcal{D}_{C_{j,l}} \times |v_{\mathcal{A}_i}|^{-1} + t^{C_j} + t^{C_l}$$
$$\forall \Re_{\mathcal{A}_i}, \exists \mathcal{T}_{\mathcal{A}_i}^M = \sum_{\substack{j=0 \\ l \neq j}}^{n} \left(\alpha \times \min\left(\mathcal{D}_{C_{j,l}}, 1\right) \times \mathcal{T}_{C_{j,l}}\right), \alpha \in \{0,1\}$$

$$Cost_{\Re_{\mathcal{A}_i}} = \lambda_1^p |\mathcal{T}_{\mathcal{A}_i}^M - \mathcal{T}_{\mathcal{A}_i}^{\nabla}| + \lambda_2^p \left(\sum_{j=1}^n \alpha C_j \times \rho^{C_j}\right)^{-1} + \lambda_3^p * \gamma_{\Re_{\mathcal{A}_i}} \quad (9)$$
$$\gamma_{\Re_{\mathcal{A}_i}} = \varepsilon \times \max(0; \mathcal{T}_{\mathcal{A}_i}^M - \mathcal{T}_{\mathcal{A}_i}^{\nabla})$$
$$Cost_{total} = \sum_{i=1}^k Cost_{\Re_{\mathcal{A}_i}}$$

where $\mathcal{T}_{\mathcal{A}_i}^M$ is the mission time completed by $\mathcal{A}_i$, $\mathcal{T}_{\mathcal{A}_i}^{\nabla}$ is the total battery time for $\mathcal{A}_i$; $S_{xyz}^{s_i}$ and $S_{xyz}^{G_i}$ correspond to the start and goal stations for $\mathcal{A}_i$, $\mathcal{D}_{C_{j,l}}$ is the distance from COTS in $C_j$ to $C_l$ and $\Re_{\mathcal{A}_i}$ is the route travelled by $\mathcal{A}_i$ with ground referenced velocity of $|v_{\mathcal{A}_i}|$. The $\alpha$ is the selection variable to show selected COTS spots in the network, while each COTS spot like $C_j$ is weighted in advance by a priority value of $\rho^{C_j}$. The total injected COTS in a mission should be maximized, and the mission time should approach the total available time for $\mathcal{A}_i$, which is represented by $Cost_{\Re_{\mathcal{A}_i}}$. The distribution pattern changes by time as the vehicles eradicate the COTS. $\gamma_{\Re_{\mathcal{A}_i}}$ is the time overdue violation to guarantee on-time completion of mission before $\mathcal{A}_i$ runs out of battery and $\varepsilon$ is a coefficient denoting the impact of violation in cost calculation. Finally, $\lambda_i^p$ ($i = 1,2,3$) represents weighting factors to balance/ highlight the corresponding mission terms used in the cost function (9). These weighting factors are tuned based on the importance of the corresponding terms, for example the magnitude difference of mission and battery time, according to the COTS eradication mission. In this particular mission, the first and third terms in the cost function (9) are of more importance for the designer and mission.

## III. Heuristic Fleet Cooperation (HFC)

Considering the key requirements for solving the constrained MRTA (C-MRTA) problem, a Heuristic Fleet Cooperation (HFC) algorithm as a population-based evolutionary approach is presented for the first time in this work. This method comprises four operators of *Clustering*, *Ordering*, *Screening*, and *Cooperation* specifically designed for solving the C-MRTA problem.

In the first stage, the proposed HFC algorithm naturally uses an automatic subdivision mechanism through the *clustering* process to categorize the most similar tasks in groups of confined areas. In the second stage, the solutions are iteratively evolved through the *ordering* process until the shortest route with exclusive set of tasks is generated for each cluster, where no tasks remain unattended. In the third stage, the *screening* mechanism effectively discards the most distant less priority tasks to fit each individual route to the defined time constraint that is the battery life for each AUV. In this way, the algorithm guarantees completing maximum possible highest priority tasks for each vehicle in the given time threshold. Ultimately, in the fourth stage, the *cooperation* mechanism facilitates each vehicle to effectively use its residual time for assisting the other vehicles after completing its own tasks that leads to maximum use of all vehicles' batty life. The algorithm's control parameters can be adjusted iteratively which enhances the convergence rate of the algorithm. The proposed method is capable of rapid prototyping of multiple optimal solutions for efficient grouping and accomplishment of the existing tasks spread over a large operation area. The detailed mechanism of the algorithm is explained in the following steps.

### 1) Clustering Operator

In this section, the initial population is generated where each individual comprises a random sequence of tasks with uniform probability. The solution space (task sequences) is identical for each cluster and improves iteratively through the evolution operators of ordering and screening. Now let us assume $\mathscr{k}$ number of AUVs to be deployed in the operating field (given by (10)). The working environment should be divided into $c=\mathscr{k}$ exclusive groups of tasks to avoid multi-vehicle mission overlap. The K-means and Fuzzy C-means (FCM) clustering methods are utilized in this study to effectively and reasonably divide the tasks between the set of AUVs.

$$\forall \mathcal{A}_i \in \langle \mathcal{A}_1, \dots, \mathcal{A}_{\mathscr{k}} \rangle, \exists\, c_j \in \langle c_1, \dots, c_c \rangle \tag{10}$$

Here the task data is partitioned into $c$ clusters defined by $c_j \in \{c_1, \dots, c_c\}$. The FCM algorithm attempts to partition a finite collection of $n$ elements $\partial = \{\partial_1, \dots, \partial_n\}$ into a collection of $c$ fuzzy clusters with respect to the given criteria. The FCM aims to minimize an objective function (11):

$$\operatorname*{argmin}_{C} \sum_{i=1}^{n} \sum_{j=1}^{c} w_{ij}^m \|\partial_i - c_j\|^2$$
$$w_{ij} = \frac{1}{\sum_{k=1}^{c} \left( \frac{\|\partial_i - c_j\|}{\|\partial_i - c_k\|} \right)^{\frac{2}{m-1}}} \tag{11}$$

Given a finite set of data, the algorithm returns a list of $c$ cluster centers and a partition matrix $W = w_{i,j} \in [0,1]$, where each element, $w_{i,j}$, refers the degree to which element, $\partial_i$, belongs to cluster $c_j$ ($w_{i,j}$ is also called membership value). The $m \geq 1 \in R$ is a fuzzifier to determine the level of cluster fuzziness. The FCM method offers number of $c$ membership values to each task; therefore, each task belongs to all the $c$ clusters, but in different degrees of membership value $w_{ij}$. The membership of a task to a cluster depends on its distance from the center of the clusters.

In K-means method, Lloyd's algorithm [21] is applied to determine the center of the clusters, and each task belongs to the closest cluster center in the environment. Clusters are refined iteratively and converge when a saturation phase emerges where there is no further chance for changes in assignment of the clusters. The K-means clustering also attempts to minimize a squared error function as an objective function defined in (11) without using the membership values $w_{ij}$ and the fuzzifier $m$. Figure 4 illustrates the performance of different clustering approaches in space decomposition for a team of three AUVs.

As shown in Fig.4, 90 tasks are randomly distributed in a non-uniform environment, and three methods have been applied to divide the tasks into three groups. Fig.4 (a) shows the performance of K-means method, where all the three clusters are completely separated. In Fig.4 (b), the FCM method is applied and maximum membership value is the selection criterion for allocating each task to its cluster. In this method, the border areas between clusters are not strictly determined and it is likely to give rise to overlap among the clusters. In Fig.4 (c), the roulette wheel selection method is applied to choose the right cluster for each task. In this method, each cluster has the chance to pick any of the tasks, but they are more likely to embrace the tasks with higher membership value. As seen in the figure, the clusters are still distinguishable, but many intrusions are appeared in the result. However, if the density of the tasks is non-uniformly distributed in the environment, this method can contribute to creating quantity balance among clusters.

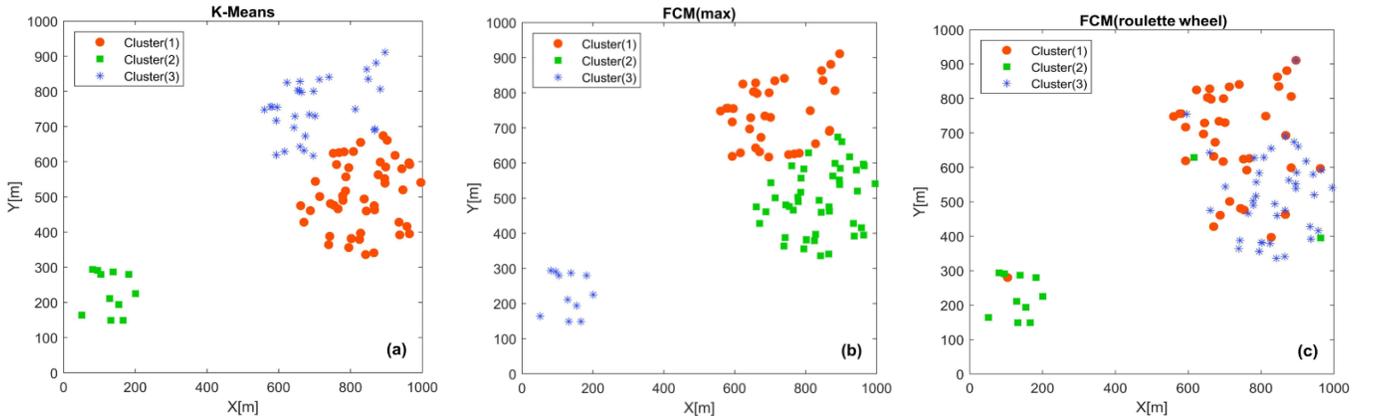

Fig. 4. Clustering the COTs distribution using the K-means method (a); FCM method with Max operator (b); and FCM with the roulette wheel operator (c).

Considering the simulation results, illustrated in Fig.4, all three mentioned approaches are suitable to perform the initialization step. Once the initial population is identically generated for each cluster, the solutions pass through the ordering process (as an evolution operator) to include all the tasks in the best possible order with respect to their locations on the map and their priority value. This results in the shortest identical route for each cluster regardless of any time restriction.

### 2) Ordering Operator

The ordering operator plays a pivotal role in the evolution process of route planning. In this stage, for each cluster, the catching order of the tasks is changed with the aim of minimizing the route length and mission time. The process starts with randomly selecting a number of feasible solution vectors from the initial population of each cluster. A cost function is defined by (9) to validate the solutions' quality during the evolution process. To change the placement sequence of the tasks in the routes, three conventional mechanisms namely Swap, Insertion, and Reversion are applied [22], [23]. Fig. 5 shows an example of the three mentioned methods used for the ordering stage.

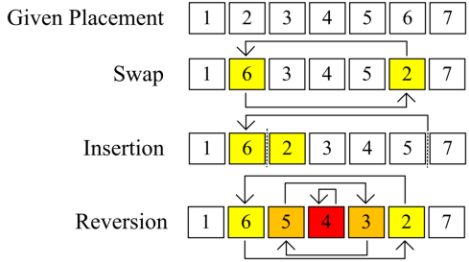

Fig.5 Mechanism of ordering operation.

Assuming two tasks of #2 and #6 are selected randomly. In the swap method, the order of the two selected tasks are replaced with each other, while in the insertion method, task number 6 has changed its place to before the task number 2. As illustrated in the figure, to perform the reversion technique, all selected tasks should be replaced two by two. Over the iterations, each individual in the population is ordered by a randomly selected mechanism and gets assessed by the defined cost function. In a case that the new solution refers to a shorter route, it replaces the prior individual in the population, and the best fitted solutions are transferred to next generation, and the rest will be omitted. Therefore, the average fitness of the new population generated by the ordering operator is improved iteratively relying on the cost evaluation process and adaptive search nature of the HFC. The process continues until the shortest set of routes for each cluster is generated to cover all the existing tasks in the best possible priority order. Ultimately, the best solutions, provided by ordering process, are fed into screening stage and get further adjusted with respect to the time constraint. This process is explained in Algorithm (1).

### 3) Screening Operator

The screening operation is a mechanism designed in this study to effectively eliminate the less important tasks out of the AUV's route to assure accurate mission timing. The pseudocode in Algorithm (2) describes the mechanism of screening operator. Let us consider the $T_{available}$ as the maximum operation time assigned for the AUVs' mission (see (12)); if the time is not sufficient to cover all the tasks in a cluster, it leads to violation of time threshold, and some tasks should be abandoned (to eliminate the violation value).

$$T_{available} = \sum_{j=1}^{k} \mathcal{T}_{\mathcal{A}_j}^{\nabla}$$
$$T_{diff}(j) = \mathcal{T}_{\mathcal{A}_j}^{\nabla} - \mathcal{T}_{\Re(j)}^{\mathcal{M}}$$
(12)

where, $\mathcal{T}_{\mathcal{A}_j}^{\nabla}$ is the total battery time for $\mathcal{A}_j$, $\mathcal{T}_{\mathcal{A}_j}^{\mathcal{M}}$ is the mission time completed by $\mathcal{A}_j$, and $T_{diff}$ is the residual battery time. The vehicles should complete the highest priority tasks in such a way that the largest possible area is covered, and the least number of far-off tasks (distant ones) need to be abandoned (to meet the defined time constraint). In other words, the screening approach should create a balance between two factors: the maximum width (expanse) of covering area, and the minimum number of tasks to be abandoned. The screening mechanism is not depended in the position of the tasks and their distance from the initial point; therefore, the tasks are uniformly abandoned form the whole area, and distant parts of the cluster will not be untouched. As a result, the widest possible area is covered by the deployed vehicles.

As shown in Algorithm (2), in each iteration a certain number of individuals are randomly selected to be screened. The cost

| Algorithm (1) – *Pseudocode of Ordering* | |
|---|---|
| **Input:** $\Re_{\mathcal{A}_j}^N: \langle S^{s_j}, ..., C_i, ..., S^{G_j} \rangle$ | //Take number of N input individual as "route" for each vehicle $\mathcal{A}_j$, where $S^{s_j}, S^{G_j}$ are start and goal stations |
| $n = \text{size}(\Re_{\mathcal{A}_j} = \Re_j)$, | //Number of the tasks in route $\Re_{\mathcal{A}_j}$ |
| **For** $j = 1$ to $N$ | // For all routes in population |
|   **For** $i = 1$ to $n$ | |
|     $L(ind(i)) = \sum_{j=1}^{n} \text{dist}(ind(i,j))$ | // Calculate the Length of the routes |
|     $r = \text{rand}(1,2,3)$ | // Select a random ordering method |
|     **if** $r = 1$ | |
|       $ind_o(i) = \text{doSwap}(ind(i))$ | // Do Swap |
|     **else-if** $r = 2$ | |
|       $ind_o(i) = \text{doInsertion}(ind(i))$ | // Do Insertion |
|     **else-if** $r = 3$ | |
|       $ind_o(i) = \text{doReversion}(ind(i))$ | // Do Reversion |
|     **end if** | |
|     $L(ind_o(i)) = \sum_{j=1}^{n} \text{dist}(ind_o(i,j))$ | // Calculate the length of the ordered route |
|     **if** $L(ind_o(i)) \leq L(ind(i))$ | |
|       $ind(i) = ind_o(i)$ | // Replace the route with the ordered one |
|     **end if** | |
|   **end For** | |
| **end For** | |
| **Output:** ordered individuals | |

value of the candidate is sequentially evaluated in the absence of only one task. In the output of the screen, the obtained cost values are compared, and the lowest cost value decides which task to be abandoned. The violation is gradually eliminated, and the solution quality is enhanced through the iterations. It must be noted that the number of candidate ($nSc$) should be selected carefully as the high level of screening rate may cause immature convergence of the algorithm. It is experientially discovered that the number of screening candidates should not be more than one percent of the population size, i.e., $nSc \leq 0 \cdot 01N$.

*4) Cooperation Operator*

In highly non-uniform task distribution environments, the size of clusters is usually non-uniform as well. In this case, some agents abandon several tasks to accomplish their mission without violating the time threshold constraint. On the other hand, in the smaller clusters, the AUV may complete all the tasks while still some battery time is left; this can be used for completing the remaining tasks in other clusters. The main role of cooperation operator is to employ idle AUVs to accomplish the remaining tasks of any nearby cluster. This contributes to equitable use of time among the AUVs and enhances the productivity. This process should be repeated until the unused time is consumed. It should be noted that during the route planning process, the travel time from the last accomplished task to the pre-determined rendezvous point should be taken into account. Algorithm (3) illustrates the pseudocode of cooperation stage. As expressed in the algorithm, by checking the remaining time of the AUVs, the idle vehicles are determined. Each idle AUV receives the position information of the abandoned tasks from the other AUVs and after calculating the distance between the last task of the route and all the abandoned tasks, overtakes the closest one into its route (after overtaking the new task the ordering operator re-orders the cluster in presence of the new task). The cooperation is repeated until the remaining time is consumed. Equation 13 gives the distance of the $i'^{th}$ abandoned task from the latest task in the route.

$$\mathcal{D}_{C_{i'}} = \sqrt{\left(C_{i',x} - C_{end,x}^{\mathcal{A}_j}\right)^2 + \left(C_{i',y} - C_{end,y}^{\mathcal{A}_j}\right)^2 + \left(C_{i',z} - C_{end,z}^{\mathcal{A}_j}\right)^2} \quad (13)$$

where $C_{i',x}, C_{i',y}, C_{i',z}$ are the coordinates of the $i^{th}$ abandoned task, and ($C_{end,x}^{\mathcal{A}_j}, C_{end,y}^{\mathcal{A}_j}, C_{end,z}^{\mathcal{A}_j}$) are the coordinates of the last executed task by the idle AUV $\mathcal{A}_j$. The distance should be obtained for the all abandoned tasks. The described sections are iteratively applied to the population until the optimal solution is achieved. Algorithm (4) is the general pseudocode of the proposed HFC algorithm.

---

**Algorithm (2) – *Pseudocode of Screening Stage***

**Input:** the individual to be screened

```
For j = 1 to k
    ℜ_{A_j}^N: ⟨S^{s_j}, ..., C_i, ..., S^{G_j}⟩        //Take number of N input individual as "route" for each
                                                      vehicle A_j, where S^{s_j}, S^{G_j} are start and goal stations
    n' = size (ℜ_{A_j} = ℜ_j),                        //Number of the tasks in route ℜ_{A_j}
    l = {1,2, ... n}                                  //Allocated tasks index
    For i = 1 to n                                    //For Each task with index of i
        While 𝒯_{A_j}^M < 𝒯_{A_j}^∇                   // check the time violation
            ℜ_j \ { l } = {C_i: C_i ∈ ℜ_j, ~ (C_i ∈ {l})}   //exclude task C_i from route ℜ_j
            mask = ones (1, n)                        //Create a 1-by-n neutral matrix as mask
            mask(i) = 0                               //Disable the i^th element in mask
            SR(C_i) = mask × ℜ_j                      //Disable the C_i by applying the mask
            Cost_{ℜ_j}(SR(C_i))                       //Calculate the cost in the absence of the C_i
            𝒯_{A_j}^M = (𝒟_{ℜ_j} - 𝒟_{ℜ_j}(SR(C_i))) / |v_{A_i}| - t^{C_i}   // calculate the mission time 𝒯_{A_j}^M in absence of task C_i
        end
        if Cost_{ℜ_j}(C_i) > Cost_{ℜ_j}(SR(C_i))      //Find the order number of the min cost
            ℜ = ℜ_j(SR(C_i))                          // Excluding task C_i from ℜ_j
        else-if Cost_{ℜ_j}(C_i) < Cost_{ℜ_j}(SR(C_i))
            ℜ = ℜ_j(C_i)                              // including task C_i in ℜ_j
        end if
    end For
    ScreenedRoute = ℜ                                 //Find the best screened route
end For
```

**Output:** screened individuals

---

**Algorithm (3) – *Pseudocode of Cooperation Stage***

- Get the label of the abandoned tasks $q = \{C_{i'}, ..., C_{i''}\}$
- Get the position of the abandoned tasks: $\{C_{i',xyz}, ..., C_{i'',xyz}\}$
- Find the idle vehicle: $\mathcal{A}_j = \mathcal{A}_{idle}$
- Get the absolute velocity of the idle vehicle $|v_{\mathcal{A}_j}|$

```
ℜ_j = ℜ_{A_idle}                                      // Take the idle vehicle's route as "route"
While 𝒯_{A_j}^M < 𝒯_{A_j}^∇                           𝒯_{A_i}^∇ is the total battery time for A_i
    n' = size (q)                                     // Number of abandoned tasks.
    For j' = 1 to n'
        l' = ind{q}                                   // Abandoned tasks index
        For i' ∈ l' do                                // For each abandoned task
            get C_{i',xyz}                            // Get the position of the abandoned tasks
            𝒟_{C_{i'}} = dist(C_{i',xyz}, C_{end,xyz}^{A_j})   // Calculate the distance of the route's endpoint
                                                      C_{end}^{A_j} from the abandoned tasks C_{i'}
        end For
        ∀i' ∈ l'; C_{i'}(best) = min(𝒟_{C_{i'}})      // Find the order number of the minimum distance.
        𝒯_{C_{i'(best)}} = Σ_{j'=1}^{n'} (𝒟_{C_{i'(best)}} × |v_{A_j}|^{-1}) + t^{C_{i'(best)}}   // 𝒯_{C_{i'(best)}} is the time for completing the abounded tasks of C_{i'}(best)
        ℜ_{j'} = [ℜ_j + C_{i'}(best)]                 // Overtake the nearest abandoned task
    end For                                           // repeat the process
    ∀ℜ_{j'}; ∃ 𝒯_{A_j}^M = (𝒯_{ℜ_j}) + 𝒯_{C_{i'(best)}}   // calculate the mission time(𝒯_{A_j}^M) for ℜ_{A_j}
end For
```

**Output:** BestCooperation = [ℜ_{j'}]

| **Algorithm (4) – *Pseudocode of HFC algorithm*** |
|---|
| *Inputs:* |
|    Population size ($N$), Population Index ($P$), |
|    Maximum iteration ($t$), Number of vehicles ($\hbar$), |
|    Maximum available time ($\mathcal{T}_{\mathcal{A}_j}^{\nabla}$) for vehicle $\mathcal{A}_j$, |
|    Number of screened individuals ($nSc$), |
|    Number of tasks in a route ($n$) |
| 1.   *//Initialization* |
| 2.     define $tasks$: $\langle \mathcal{C}_{i,xyz}\rangle; i \in \{1,\ldots,n\}$ |
| 3.     **For** $i = 1$ to $N$ |
| 4.       $ind(i) = \text{cluster}(tasks, \hbar)$ |
| 5.       **For** $j = 1$ to $\hbar$ |
| 6.         $\text{get}(ind(i,j))$ |
| 7.         $Cost_{\Re}(i,j) = \text{cost}(ind(i,j))$ |
| 8.       **end For** |
| 9.       $\Re(i,j) = ind(i,j)$ |
| 10.     **end For** |
| 11.   *// HFC main loop* |
| 12.     **For** $k = 1$ to $t$ |
| 13.       *//Ordering* |
| 14.       **For** $i = 1$ to $N$ |
| 15.         **For** $j = 1$ to $\hbar$ |
| 16.           $\Re_{ord}(i,j) = \text{order}(\Re(i,j))$ |
| 17.           $Cost_{ord}(i,j) = Cost(\Re_{ord}(i,j))$ |
| 18.           **if** $Cost_{ord}(i,j) < Cost_{\Re}(i,j)$ |
| 19.             $\Re(i,j) = \Re_{ord}(i,j)$ |
| 20.           **end if** |
| 21.         **end For** |
| 22.       **end For** |
| 23.       Output ordered individuals $\Re_{ord}$ |
| 24.       *//Screening* |
| 25.       $Sc = \text{randomSample}(\Re_{ord}, nSc)$ |
| 26.       **For** $l = 1$ to $Sc$ |
| 27.         **For** $j = 1$ to $\hbar$ |
| 28.           **if** $\mathcal{T}_{\mathcal{A}_j}^{\mathcal{M}} > \mathcal{T}_{\mathcal{A}_j}^{\nabla}$ |
| 29.             $\Re_{scr}(l,j) = \text{screen}(\Re_{ord}(l,j))$ |
| 30.             **if** $Cost_{\Re_j}(\mathcal{C}_i) > Cost_{\Re_j}(SR(\mathcal{C}_i))$ |
| 31.               $\Re_j = \Re_{scr}$ |
| 32.             **else-if** $Cost_{\Re_j}(\mathcal{C}_i) < Cost_{\Re_j}(SR(\mathcal{C}_i))$ |
| 33.               $\Re = \Re_{ord}$ |
| 34.             **end if** |
| 35.           **end if** |
| 36.         **end For** |
| 37.       **end For** |
| 38.       Output the screened individuals $\{\Re\}$ |
| 39.       *//Cooperation* |
| 40.       **For** $j = 1$ to $\hbar$ |
| 41.         $T_{diff}(j) = \mathcal{T}_{\mathcal{A}_j}^{\nabla} - \mathcal{T}_{\Re(j)}^{\mathcal{M}}$ |
| 42.         **if** $T_{diff}(j) > 0$ |
| 43.           $\Re'(j) = \text{cooperate}(\Re(j))$ |
| 44.           $\Re_{best}(j) = \text{order}(\Re'(j))$ |
| 45.         **end if** |
| 46.       **end For** |
| 47.   |
| 48.     **end For** |
| *Output:* |
|    The shortest route $\Re$ for each $\mathcal{A}_{\hbar}$ with minimum $T_{diff}$ and best order of maximum completed tasks. |

According to Algorithm (4), the HFC has a loop corresponding to the number of the vehicles ($\hbar$). Hence, analysis of the algorithm itself may result in a judgment that it converges with the complexity of $O(\hbar Nt)$. However, considering the complexity of the MRTA problem over all possible instances, the complexity of HFC also relies on the implementation of the evolution operators (*Ordering*, *Screening*, and *Cooperation*), design and encoding of the individuals in the population, and certainly the complexity of the cost function that can significantly impact the convergence speed of the HFC. Hence, given the possible choices for evolution operators, at the extreme condition, the algorithm complexity is obtained as follows:

$$O\left(\hbar Nnt\left(2 + \frac{1}{n}\right)\right)$$
$$= O\left(\left(\overbrace{\hbar Nn}^{\text{Ordering}} + \overbrace{\hbar Nn}^{\text{Screening}} + \overbrace{\hbar N}^{\text{Cooperation}}\right) \times t\right) \quad (14)$$

where, $N$ is the population size and $n$ the size of the individuals (solution vectors). The cost function complexity is ignored in this computation as it depends on the application.

## IV. RESULTS AND DISCUSSIONS

The AUVs' mission planner in this research, uses a priori information of COTS distributions (described in Section II (A)), maximum operation time, and battery capacity of each vehicle to compute the most appropriate order of tasks (from beginning toward the destination). Perception of the operating field is achieved via the AUVS' navigation aids such as Horizontal Acoustic Doppler Current Profiler (H-ADCP) and Doppler Velocity Logger (DVL) and their situational awareness modules. Information sharing and exchange between the AUVs are carried out via acoustic commination using DUSBLs.

In this study, all computations were performed on a desktop PC with an Intel i7 3.20 GHz quad-core processor in MATLAB®2019a. In the subsequent sections, different mission scenarios are defined, and the performance of the proposed algorithm in eradicating the maximum number of COTS is evaluated in detail.

### A. Qualitative assessment of HFC-based mission planning

To evaluate the performance productivity of the proposed algorithm on multi-vehicle cooperative task assignment and completion, the following modes are defined:

- ***i.** Non-Cooperative* mode 1 (NCM1): where only the ordering operator is switched on while screening and cooperation operators remain off;

- ***ii.** Non-Cooperative* mode 2 (NCM2): where the ordering and screening operators are switched on and the cooperation operator remains off;

- ***iii.** Cooperative* mode (CM): where the ordering, screening and cooperation operators are switched on;

In the first scenario, the HFC-based mission planner's performance is investigated in the NCM1 mode, where the vehicles do not communicate to each other; however, the catching order of the tasks for each cluster is changed with the aim of minimizing the route length and mission time. In this mode, each vehicle individually plans its mission regardless of the others' mission. The time constraint is not considered in NCM1 mode, and each vehicle aims at completing all the existing tasks in the assigned cluster taking the shortest route and minimizing the route time. Consequently, the expectation is to have large time violations for this scenario.

The second scenario is the augmented version of the first one in which the screening operator is also enabled to identify and decide the tasks that should be abandoned, while considering the time threshold constraint (as explained in Algorithm (2)). However, this scenario is *non-cooperative* and the main drawback of this mode is that if any vehicle completes its mission with considerable remaining time, it cannot use this time to help other vehicles in completing their tasks.

On the other hand, the CM facilitates the vehicles to communicate with each other to cover battery restrictions and to cover multiple missions, while the vehicles managing the endurance time to handle more COTs injection tasks. In this case, if any vehicle gets discarded from the team (abort its mission for any reason) or runs out of battery, the closest vehicle(s) with the most similar configurations undertake(s) the incomplete mission or the mission is divided between the operational vehicles and the planner module re-plans a new mission scenario for them. This leverages mission coverage range, areas, system resilience, and risk management.

To investigate the performance evaluation of the proposed algorithm, three homogeneous COTSbot AUVs are deployed to a coastal environment, while all three vehicles are configured with the same setting. The battery life-time threshold for each vehicle is equally set to $3.6 \times 10^3$ (*sec*), and the vehicles are assumed to be operated with a constant average velocity of 1 m/s.

Ninety tasks (Starfish spots) are non-uniformly distributed in the area of (1000×1000) $m^2$. The injection time (task completion time) for each COTS spot is 90 *sec*. The AUVs start their mission from the same initial position, and after the completion of the mission they should arrive at the rendezvous point (a priori known coordinate).

Figure 6 (*a*-1, *b*-1, *c*-1) shows a contour map of the operating field, including the tasks, the route lines, the start point and also the rendezvous point. The colormap represents the density of the COTS covered area while the probability of distribution ranges from 0 to 1 (with least to most probablity of Starfish mass in the area). In this figure, the lighter colors, leaned to yellow, correspond to the higher intensity COTS areas and the darkest parts, indicated by black color, are the shore lands. Having a closer look at the figure, the distribution of tasks is concentrated around the deeper spots denoted by lighter colors, and shallow areas are out of tasks.

In order to demonstrate the effect of the algorithm's operators, the operators are enabled step by step, and their contributions are displayed in different contour maps. As demonstrated in Fig.6, cluster (1) and (3) encircle the higher number of tasks (the 2$^{nd}$ cluster encapsulates fewer number of tasks). We use this difference to show how cooperative mode enables vehicles to undertake others' tasks in their spare time.

In the first scenario (NCM1) given by Fig.6 (*a*-1) and (*a*-2), the time constraint is not taken into account, and the goal is to find the shortest route through the tasks. In this scenario, the main purpose of ordering operator is to maximize the number of completed tasks regardless of time threshold (that is satisfied considering the pattern of completed tasks as shown in Fig.6 (*a*-2)).

In the second scenario given by Fig.6 (*b*-1) and (*b*-2), NCM2 is applied in which both screening and ordering operators are enabled to maximize the completed tasks and minimize the route length while taking the time constraint into account. As shown in Fig.6 (*b*-2), some tasks are abandoned by the 1$^{st}$ and 3$^{rd}$ AUVs due to the applied time threshold. However, the 2$^{nd}$ AUV accomplishes a few numbers of tasks encapsulated in a smaller cluster; in addition, all three AUVs arrive rendezvous point without violating the time threshold. Although the time threshold is not violated by the vehicles, the 2$^{nd}$ vehicle could effectively use its residual time for assisting the other vehicles after completing its own tasks, which is addressed using the cooperation operator. The CM is applied in the third scenario and the complete performance of the HFC algorithm is illustrated by Fig.6 (*c*-1) and (*c*-2), where the ordering, screening and cooperation operators are enabled. Considering the given task completion pattern in the CM (see Fig.6 (*c*-2)), the 2$^{nd}$ vehicle successfully completes all the existing tasks in cluster (2) in the given time and devotes its remained time to cooperate with the 3$^{rd}$ vehicle for accomplishing the abandoned tasks in cluster (3). In contrast, it is apparent that there is no such a cooperation in the first two scenarios as all vehicles just concentrated on their assigned cluster without being aware of other vehicles' operation detail. This is a good indication of how cooperation impacts the task completion performance, while the vehicles communicate and update each other with their completed tasks. Ultimately, all three vehicles terminate their mission in the rendezvous point.

Figure 7 demonstrates the algorithm performance for all three mentioned scenarios based on the mission cost metric which is a direct function of the number of completed tasks and the time violation. It is noteworthy to mention that since the evaluation criterion for deciding the best solution is the total cost of the three vehicles, in some iterations one vehicle may experience a slight increase in cost value comparing with the previous iterations; but the summation of the three costs is always less than the previous iteration. As shown in Fig.7 (*a*), (*b*), the total cost follows a decreasing trend and settles on about $2.27 \times 10^5$ after 150 iterations. Even though the number of completed tasks in the first scenario is high, the screening mode is off which means the planner is not able to well manage the time violation.

Therefore, mission planning cost in the first scenario (Fig.7 (*a*)) is considerably higher than the other two scenarios.

In the second scenario, the screening operator is enabled along with the ordering operation, meaning that the planner tends to maximize the number of completed tasks and minimize the route length, while taking time constraint into account. The time constraint enforces the vehicles to abandon some of the tasks in the crowded clusters. Considering the total cost trend in Fig. 7(*b*) the algorithm effectively suppresses the total cost value to be settled on $0.71 \times 10^5$, which is remarkably less than the produced cost in Fig.7 (*a*).

Although the total cost in Fig.7 (*b*) is significantly reduced compared to the first scenario, it is almost twice greater than the total cost in Fig.7 (*c*) (all vehicles are only responsible for their assigned clusters). It is inferred from Fig.7 (*c*) that the best performance in minimizing the cost belongs to the 3$^{rd}$ scenario ($0.38 \times 10^5$) in which the vehicles can cooperate with each other.

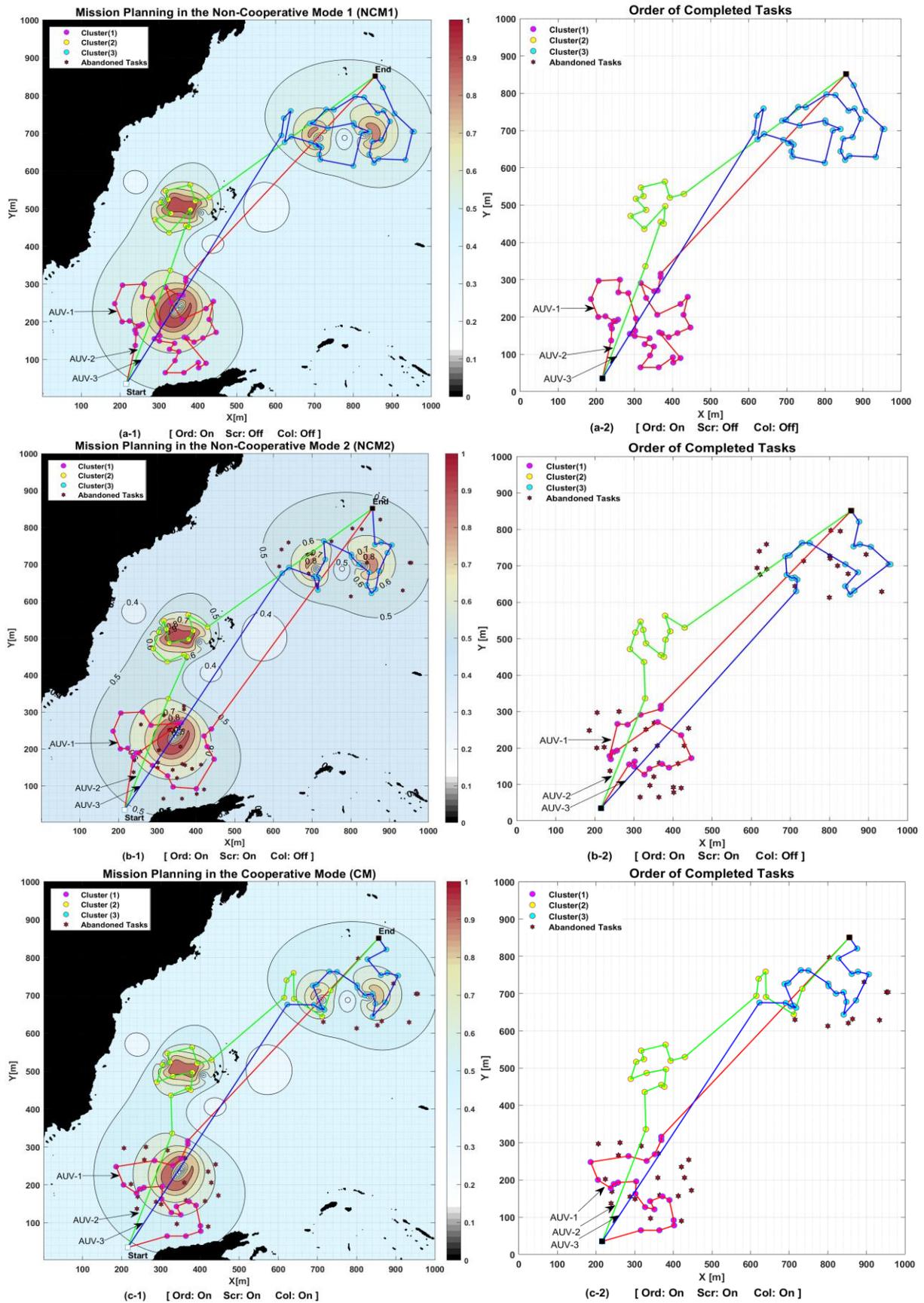

Fig.6. (*a*-1) and (*a*-2) The task completion pattern by multiple vehicles in NCM1, where ordering operator is on, screening and cooperation operators remain off;

(*b*-1) and (*b*-2) The task completion pattern in NCM2, where the ordering and screening operators are functioning but cooperation operator is disabled;

(*c*-1) and (*c*-2) CM, where the ordering, screening and cooperation operators are functioning.

The total cost and the cost of individual vehicles are reduced iteratively meaning that the population effectively converges to the optimum solution, i.e., taking the best use of mission available time.

To further assess the performance of the proposed algorithm on the given scenarios, qualitative assessment is undertaken with performance indices of time violation and time difference between battery lifetime (considered 3600 sec) and the actual operation time for each vehicle.

Figure 8 demonstrates the evolution trend of the proposed algorithm in the mentioned scenarios. Fig.8 (*a*-2) shows that the 1st and 3rd AUVs take negative values of time-difference ($T_{diff}$), and they cannot entirely eliminate the violation (referring to Fig.8 (*a*-1)); this means that the two mentioned AUVs experience the shortage of time for accomplishing their missions. Although the 2nd AUV does not violate the time limit (due to the small size of cluster (2)), the corresponding $T_{diff}$ is significantly high; this time difference could be used to reduce the load of other vehicles, and help them to eliminate their time violations.

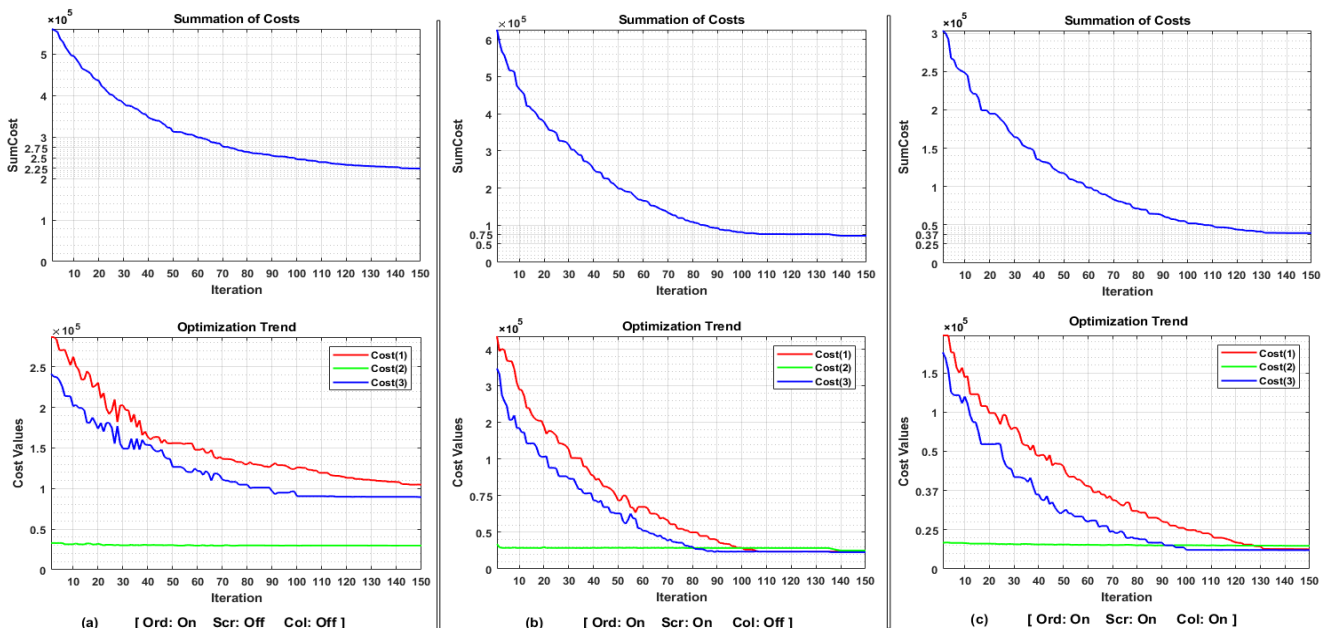

Fig.7 Mission performance in terms of total and individual vehicles' cost variation over 150 iterations where: (a) only ordering mode is functioning while screening and cooperation operators are disabled (*NCM1*); (b) ordering and screening mode are functioning but the cooperation mode is disabled (*NCM2*); (c) ordering, screening and cooperation mode are switched on (*CM*).

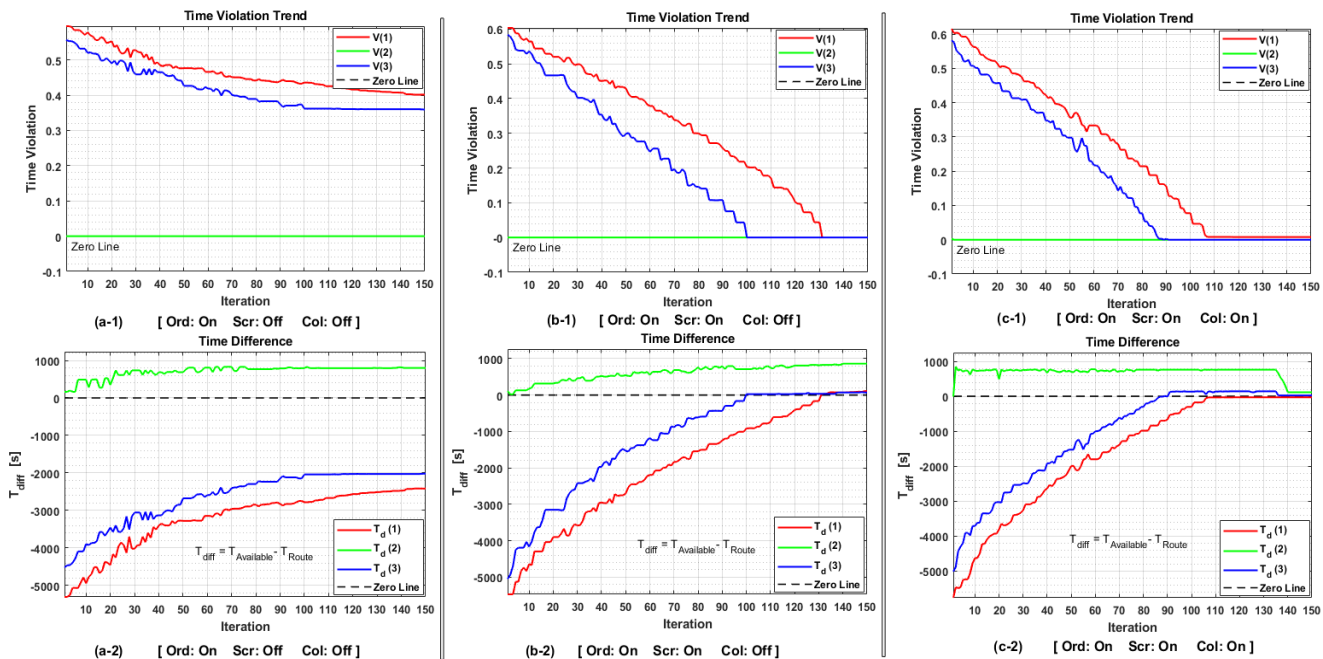

Fig.8 Mission performance in terms of individual vehicles' Time violations in three scenarios of (*a*-1) only ordering mode is functioning while screening and cooperation operators are disabled (*NCM1*); (*b*-1) ordering and screening mode are functioning but the cooperation mode is disabled (*NCM2*); (*c*-1) ordering, screening and cooperation operator are functioning (*CM*). Mission time difference ($T_{diff}$) for individual vehicles' over 150 iterations where: (*a*-2) *NCM1*; (*b*-2) *NCM2* (*c*-2) *CM*

TABLE I. NUMERICAL ASSESSMENT OF THE ALGORITHM PERFORMANCE IN NCM1, NCM2, AND CM.

| Scenarios | Active Operators | Operation Cost | Operation Duration | Time Difference | Completed Tasks NO | Violation |
|---|---|---|---|---|---|---|
| **NCM1** | Ord: ON<br>Scr: OFF<br>Col: OFF | $V_1 \cong 1.053 \times 10^5$<br>$V_2 \cong 0.317 \times 10^5$<br>$V_3 \cong 0.903 \times 10^5$ | $V_1 = 5,602.00$ (sec)<br>$V_2 = 2,794.00$ (sec)<br>$V_3 = 6,033.00$ (sec) | $V_1 = -2,002$ (sec)<br>$V_2 = +806$ (sec)<br>$V_3 = -2,433$ (sec) | $V_1 = 41$<br>$V_2 = 13$<br>$V_3 = 36$ | $V_1 = 0.4081$<br>$V_2 = 0.0000$<br>$V_3 = 0.3630$ |
| | Total= | $2.273 \times 10^5$ | 14,429 (sec) | $V_1, V_3$: Failed<br>$V_2 = +806$ (sec) | 90 / 90 | 0.7711 |
| **NMC2** | Ord: ON<br>Scr: ON<br>Col: OFF | $V_1 \cong 0.227 \times 10^5$<br>$V_2 \cong 0.271 \times 10^5$<br>$V_3 \cong 0.216 \times 10^5$ | $V_1 = 3,587.03$ (sec)<br>$V_2 = 2,782.97$ (sec)<br>$V_3 = 3,591.84$ (sec) | $V_1 = +12.97$ (sec)<br>$V_2 = +817.03$ (sec)<br>$V_3 = +08.16$ (sec) | $V_1 = 19$<br>$V_2 = 13$<br>$V_3 = 20$ | $V_1 = 0.0000$<br>$V_2 = 0.0000$<br>$V_3 = 0.0000$ |
| | Total= | $0.714 \times 10^5$ | 9,962 (sec) | $+ 838.16$ (sec) | 52 / 90 | 0.0000 |
| **CM** | Ord: ON<br>Scr: ON<br>Col: ON | $V_1 \cong 0.122 \times 10^5$<br>$V_2 \cong 0.143 \times 10^5$<br>$V_3 \cong 0.121 \times 10^5$ | $V_1 = 3,582.56$ (sec)<br>$V_2 = 3,506.93$ (sec)<br>$V_3 = 3,585.74$ (sec) | $V_1 = +17.48$ (sec)<br>$V_2 = +93.07$ (sec)<br>$V_3 = +14.26$ (sec) | $V_1 = 21$<br>$V_2 = 19$<br>$V_3 = 21$ | $V_1 = 0.0000$<br>$V_2 = 0.0000$<br>$V_3 = 0.0000$ |
| | Total= | $0.386 \times 10^5$ | 10,676 (sec) | $+ 124.81$ (sec) | 61 / 90 | 0.0000 |

Referring to Fig.8 (*b*-1) and (*b*-2), the performance of the algorithm is slightly improved due to enabling the screening operator. This results in managing the operations' time violation.

However, due to the large size of clusters (1) and (3), the vehicles are forced to ignore some of the tasks to satisfy the time constraint, and consequently to eliminate the violation. However, the 2nd vehicle ends up with a large positive time difference, which this time could be used to handle unaddressed tasks of the nearby clusters.

Figure 8 (*c*-1) and (*c*-2) demonstrate the ultimate performance of the algorithm while the cooperation operator is enabled. In this scenario, the 2nd AUV devotes its remained time to the other AUVs to cooperate in tasks' handling. The time difference in Fig.8(*c*-2) and the violation graphs in Fig.8 (*c*-1) completely endorse the performance of the proposed algorithm. As illustrated in in Fig.8 (*c*-1), the 2nd AUV does not violate the time constraint over the 150 iterations while the 1st and 3rd AUVs have experienced the high level of violation at early iterations; however, the algorithm gradually reduces the violation value and touches the zero line after 100 iterations.

Tracking the trend of cost for each vehicle in Fig.7(*c*), Fig.8(*c*-1), and Fig.8(*c*-2) reveals that the algorithm accurately enforces the solutions to approach the least cost and navigates the solutions to eliminate the violation within 150 iterations in the CM. Table I provides a detailed numerical comparison of the algorithm performance for the mentioned scenarios.

As indicated in Table I, in the NCM1, vehicles (1) and (3) terminate the mission with a large negative time difference of $V_1 = -2,002$ and $V_3 = -2,433$ (*sec*), respectively. This means that the planner in NCM1 does not consider the time restriction as the screening operator is disabled. Consequently, the mission planner in the NCM1 can meet the maximum number of tasks as no restriction confines the planner to avoid negative time difference.

Due to the smaller size of cluster (2) that includes only 13 tasks, the 2nd vehicle can complete all the 13 tasks in its mission with positive time difference of $V_2 = +806$ (*sec*), where this time can be used to take the negative load of other vehicles and reduce their violation. It should be noted that the cost value has a direct relation to the violation and therefore, the cost value for the first scenario is considerably higher than the other two modes.

In the second scenario, there is no violation, and $T_{diff}$ for all vehicles tends to have a positive value; however, the operation cost of $0.714 \times 10^5$ is still larger than the third scenario with total cost of $0.386 \times 10^5$. The reason is that the second scenario only enables ordering and screening mode helping the vehicles to individually optimize the route length and tasks quantity with no time violation. This enforces the vehicles in larger clusters to ignore most of the tasks to meet time restriction. However, the vehicles in smaller clusters can complete all their tasks in a small portion of their available time and finish their mission without considering the nearby clusters. As a result, only 52 tasks out 90 are completed which leads the planner to experience a higher cost compared to the cooperative mode.

Ultimately, Table I shows that the operation time for all vehicles in the CM is smaller than total available time, and the violation value for all solutions is equal to zero. This confirms the feasibility of the produced routes. The total $T_{diff}$ for the cooperative operation is only $+ 124.81$ (*sec*) out of total available time 10800 sec ($3 \times 3.6 \times 10^3$) for three vehicles. Obviously, the total operation cost for the collaborative mode ($0.386 \times 10^5$) is significantly less than the other two scenarios due to use of all the three operators of ordering (maximizes number of completed tasks), screening (satisfies time restriction), and cooperation (facilitates efficient mission time management), confirming the superior performance of the CM. As can been seen in the CM, the 2nd AUV completes all the 13 tasks in its cluster and allocates its remaining time to cooperate with the 3rd AUV to handle 6 more tasks; this ends up with a small time difference of $+ 93.07$ (*sec*).

Referring to the results in Table I, it is notable that the operation cost for CM experiences a decrease of 83.02% compared to NCM1 and 45.94% with respect to operation cost of the planner in NCM2. The planner in NCM1 completed 100% of the existing tasks due to having no time restriction while this number is decreased to 58% of existing tasks for NCM2 due to added time restriction. In the third scenario, 67.78 % of existing tasks are completed indicating a decline of 32.22% with respect to NCM1 due to having time restriction and improvement of 9.78% compared to NCM2 due to enabling the cooperative operator. In the sequel, the residual time for CM is minimized to 14.89% of the residual time for NCM2, which means the best use of total available time is taken in the CM to complete maximum possible number of tasks.

## B. Quantitative assessment of HFC-based mission planning

To assess the robustness and reliability of the designed cooperative mission planner, 30 execution runs are performed in a Monte Carlo simulation for each vehicle in the CM. Figs.9-12 demonstrate the quantitative performance of the cooperative mission planner in dealing with problem's space deformation and against several significant mission's metrics including mission cost, time difference ($T_{diff}$), and number of successful task completion (number of killed COTs spots). The COTs distribution and the topology of the graph is changed randomly based on a Gaussian distribution on the problem search space. The total battery life is set equally with $3.6 \times 10^3$ (*sec*) for each vehicle. The desired performance is defined based on the maximum number of COTS kills subject to the vehicles' battery capacity /mission time ($\sum_{j=1}^{k} \mathcal{T}_{\mathcal{A}_j}^{\nabla}$). Time violation is for the case that any of the AUVs goes beyond the specified time threshold, and its operation takes longer than what battery capacity allows.

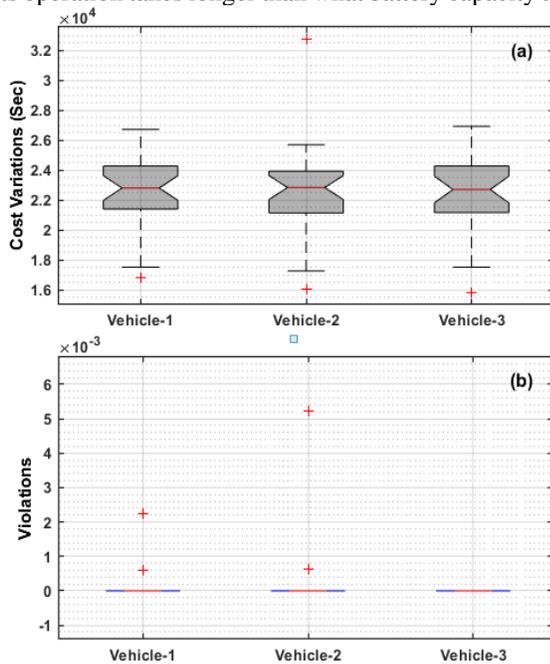

Fig.9 (a) Mission cost variation in 30 Monte Carlo simulations in dealing with space deformation; (b) time violation over the given time threshold of $3.6 \times 10^3$ (*sec*) for each vehicle in 30 experiments.

Figure 9 (*a*) shows that the median of mission cost is about 2.3 for all three vehicles with maximum range of variation about 0.5 under the Monte Carlo experiments; this reveals the robustness of the planner against variations. It should be noted that within the 30 Monte Carlo runs, there just exist four failures (13% failure rate) indicated by outliers in Fig.9 (*a*). Fig.9(*b*) shows that the mission planner is able to meet the specified constraint as the time violation for all the three vehicles in 30 missions is centralized in zero, which is a good indication of system's time management capability. The violation diagram shows how the generated solutions respects the defined restriction (restriction of mission time to available time and feasibility criteria). This means the mission planner accurately enforces the solutions for all the three vehicles to approach the least cost and manage the solutions to eliminate the violation as the range of violation variations converges to zero in all Monte Carlo experiments.

Figure 10 illustrates the results of Monte Carlo trials against performance metrics such as rout length, mission time, and number of task completion.

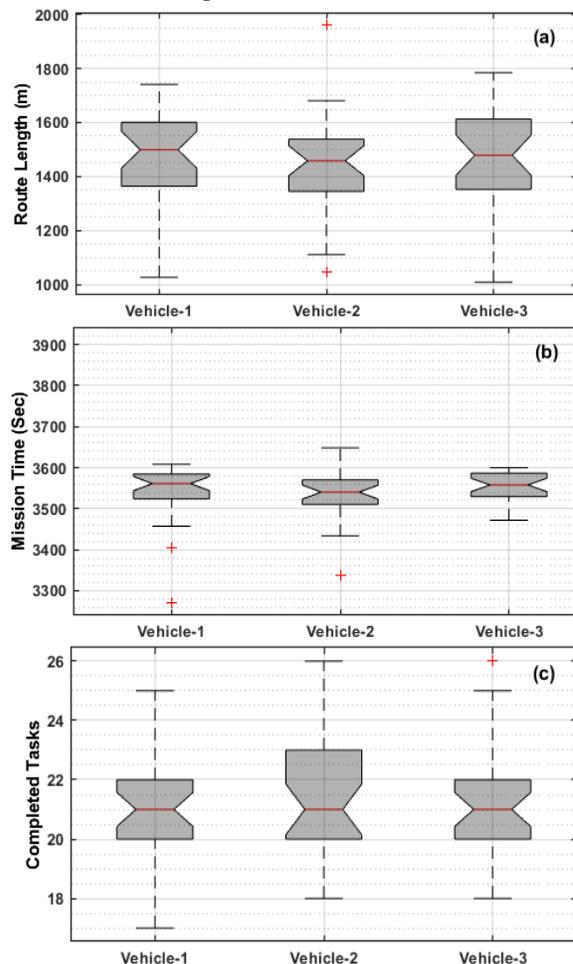

Fig.10 (a) Average distance travelled by each vehicle in 30 Monte Carlo simulations; (b) Average operation time for each vehicle; (c) Average number of completed tasks by each vehicle.

Fig.10 (*a*) shows that the minimum and maximum statistical range of traveled rout belongs to the second vehicle (~700 *m*) and third vehicle (~ 800 *m*) respectively while all three vehicles have an approximate median route length of 1.5 *km*. This is followed by the boxplots in Fig.10 (*b*) indicating the variations of the operation time for all the three vehicles is centralized in a very narrow boundary in range of $3.5 \times 10^3$ to $3.6 \times 10^3$ *sec* in all executions. This confirms that all three vehicles tend to efficiently use their maximum battery time ($3.6 \times 10^3$ *sec*) in the CM. In addition, Fig.10 (*b*) indicates that the planner robustly makes a balance between *available time* and *the actual operation time* as the average variations for all executions are lied in similar range and very close to the upper bound of $3.6 \times 10^3$ *sec*. This confirms efficient cooperation of the vehicles in handling the tasks in a way that none of them is left with considerable time difference (residual battery time). Also, Fig.10(*c*) demonstrates that the vehicles in the CM are able to accomplish the COTS killing task with 70% success rate (sum median of 63 out of 90) that is consistent with the success rate of 67% indicated in Table I. This is another indication of the algorithm robustness under dynamicity and uncertainty of the operating environment.

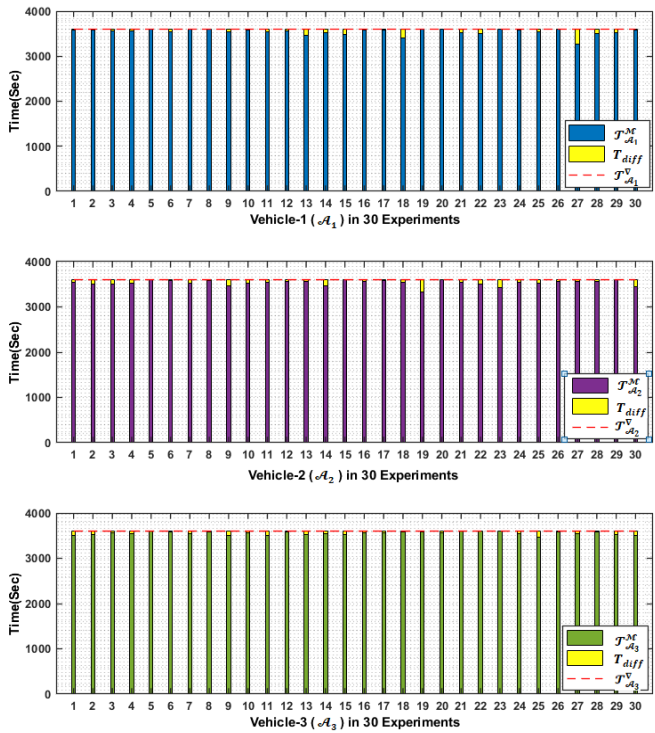

Fig.11 Time management performance for each vehicle in the CM.

Figure 11 shows time management performance analysis of the mission planner for all three vehicles under 30 Monte Carlo trials. The operation commences with the equal time constraint of $\mathcal{T}_{\mathcal{A}_i}^{\triangledown}=3.6\times10^3$ *sec* for each vehicle (presented by red dashed horizontal line in Fig.11). As shown in Fig.11, the $T_{diff}$ for all three vehicles is a very small value in all the missions indicating optimally use of mission time ($\mathcal{T}_{\mathcal{A}_i}^{\mathcal{M}}$) in CM even under variations of operating field (good indication of the planner time management robustness). Fig. 12 provides a zoom-in view of the time difference index for all three vehicles under 30 Monte Carlo trials.

As given in Fig.12, the planner reasonably manages the time as $T_{diff}$ in most of the cases tends to have a tiny positive value meaning that the vehicles accomplish their mission before running out of battery. To be more specific, the best performance belongs to experiment #5, #8 and #16 with the minimum positive $T_{diff}$. Also, in a number of experiments, negligible delays are found; for example, experiment #7 with only 8 *sec* for vehicle $V_1$, experiment #15 with 10 *sec* for $V_2$, experiment #20 with 7 *sec* for $V_2$, and experiment #26 with 8 *sec* for $V_1$, that can be ignored as they are insignificant compared to the total available time of 3600 *sec*.

## C. Comparative Study

The purpose of this section is to evaluate the performance of the proposed HFC algorithm against an existing benchmark method of the state-of-the-art. To this end, a benchmark method presented in [24] is used. The study proposed a GA-based algorithm to solve a cooperative task allocation problem for a multi-robot system. In this work, the operation space was decomposed via k-means clustering method and the GA was employed to optimize the route length of each robot in the assigned cluster. The partially mapped crossover (PMC) method [25] was utilized for implementing the cross-over term and the mutation term played a role analogous to the ordering operator of the proposed HFC.

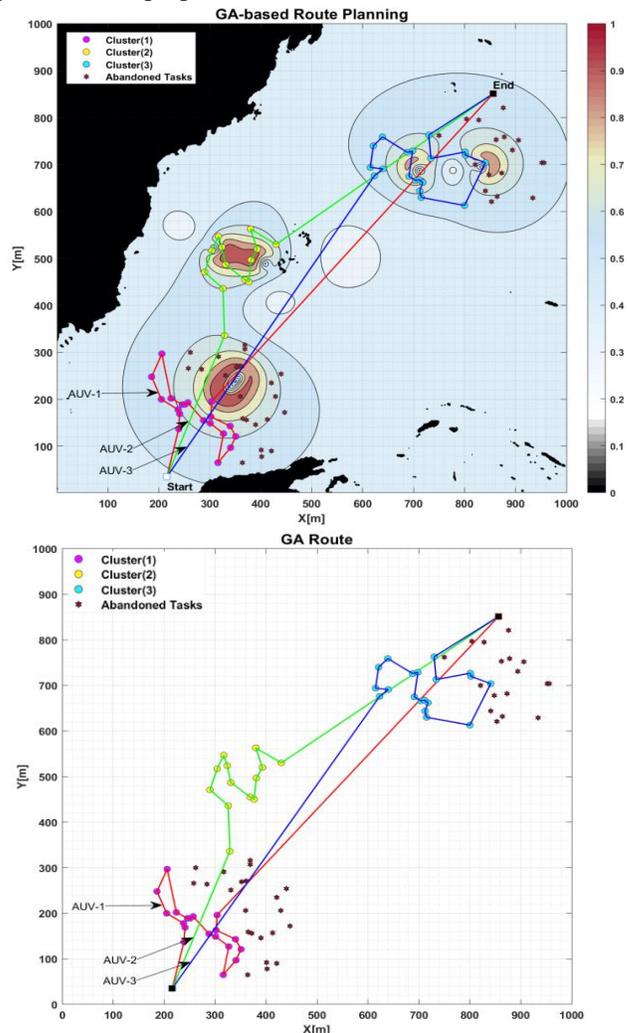

Fig.13. The task completion pattern by multiple vehicles using GA-based task assignment method.

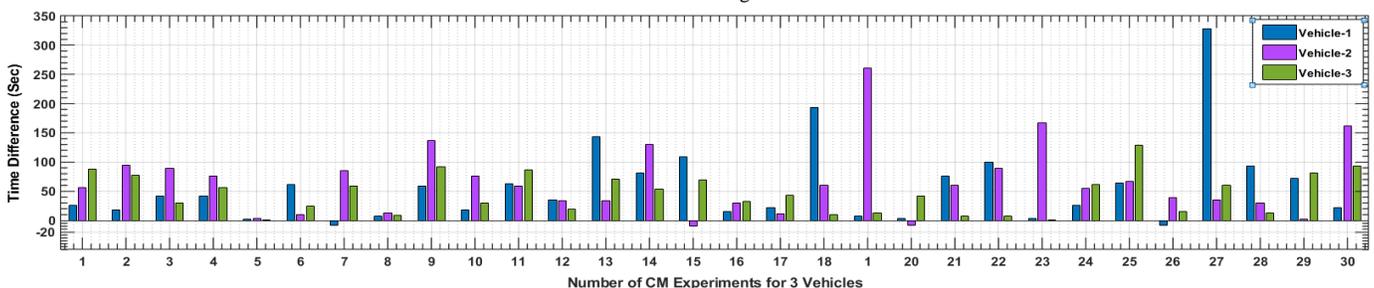

Fig.12 Variations of time difference ($T_{diff}$) between total available time of $\mathcal{T}_{\mathcal{A}_i}^{\triangledown}$ and the vehicles' actual operation time of $\mathcal{T}_{\mathcal{A}_i}^{\mathcal{M}}$.

Figure 13 demonstrate the simulation results of the GA-based task assignment method [1] for the mission scenario introduced in Section IV. As shown in the figure, the GA-based task assignment method is successfully able to provide an optimal route for each vehicle. The number of accomplished tasks is 52, which is equal to the NCM2 mode of the proposed HFC. However, in each cluster the nearby area to the start point has been completely eradicated of COTS, while the distant area from the initial position has remained untouched. In contrast, screening policy in the HFC algorithm has uniformly taken the tasks from the whole allocated area, and this contributes to reducing the density of the COTS in a wider coverage. Furthermore, the GA-based task assignment method has no mechanism to use the remained time of the smaller cluster to assist the other ones, while in cooperative mode of the HFC algorithm, by using the cooperation operator, the 2$^{nd}$ cluster has devoted its extra time to overtake more tasks from nearby clusters.

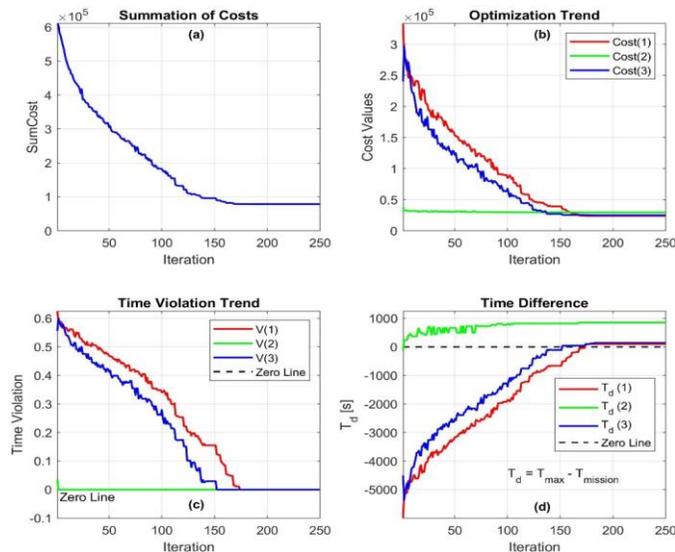

Fig.14. Mission performance criteria evaluation using GA-based task assignment method.

As shown in Fig14 (a) and (b), the algorithm has succeeded in finding the optimum solution in almost 170 iterations. Figure 14 (c) illustrates that by leaving the last ordered task in each iteration the algorithm is finally able to eliminate the violation, while the time difference factor of the 2$^{nd}$ AUV (figure (d)) has not been reduced due to the lack of cooperation mechanism in GA.

From numerical point of view, the total time difference factor is 1103 seconds with the GA-based task assignment method while in the two modes of the HFC, this parameter is 838 and 124 seconds respectively. The number of completed tasks in GA mode is 52, which is equal to NCM2 in HFC, while in the CM of HFC this number is noticeably increased to 61 tasks.

Overall, by employing the HFC algorithm in the cooperative mode, 14.7% improvement in the COTS eradicating performance, compared to the GA-based task assignment method, is achieved.

## V. CONCLUSION

This paper proposed a new approach to address the environmental problem of control of COTS in the Australia's Great Barrier Reef. To this end, a novel cooperative mission planner algorithm for a certain class of underwater vehicles namely COTSbot AUVs was developed and its performance was investigated within extensive simulation studies. The problem of COTS control was transcribed in the context of a constrained task assignment problem and cooperative operation of the AUVs was used to maximize the number of task completion. The simulation results indicate the effectiveness of this approach and the planner algorithm while comparing with the non-cooperative or individual-based AUV COTS killing operation. The robustness of the proposed planner in the cooperative operation was examined with Monte Carlo simulations by applying topological variations on the COTS clusters. The result of Monte Carlo analysis confirmed the stability and robustness of the planner in dealing with random deformation of the COTS distribution and environmental topology. Moreover, the result of the comparative study demonstrated the superior performance of the HFC algorithm against the benchmark GA-based task assignment method.

Future extension of this study includes field trials and evaluation of the planner in practice.

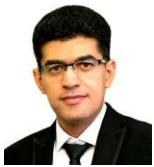
**Amin Abbasi** was graduated in Master of Control Engineering in Azad University of Khomeinishahr, Esfahan, Iran. His research field is motion control and path planning of mobile robots. He is interested in intelligent algorithms and machine learning, and their application in autonomous engineering.
 *aminabbasi.res@gmail.com*

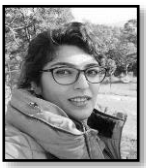
**Somaiyeh MahmoudZadeh** is Assistant Professor of Data Science at School of IT, Deakin University of Australia. She was employed as a Postdoctoral Research Fellow by Monash University since 2017 to 2018. Her research area includes Computational Intelligence, Autonomous Decision Making and Situational Awareness in Unmanned Vehicles Mission-Motion planning. *s.mahmoudzadeh@deakin.edu.au*

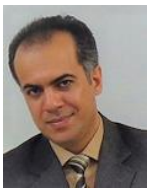
**Amirmehdi Yazdani** (S'14-M'18) received his PhD degree in Electrical-Control Engineering from Flinders University, SA, Australia, in 2017. From 2017 to 2018, he was employed as a Postdoctoral Research Associate by Flinders University. He is currently working as a Lecturer in Electrical Engineering in the College of Science, Health, Engineering and Education at Murdoch University, WA, Australia. He is serving as an Academic Chair of Engineering Technology, Electrical Power Engineering, and Renewable Energy Engineering at Murdoch University. He is also the Vice-Chair of IEEE Industrial Electronic Society, WA Chapter. His areas of research specialization are concerned with guidance and control of robotic, autonomous, and mechatronic systems, optimal control and state estimation theory, and intelligent control applications. *amirmehdi.yazdani@murdoch.edu.au*